\documentclass[conference]{IEEEtran}
\IEEEoverridecommandlockouts
\usepackage{amsmath,amssymb,amsfonts}
\usepackage{graphicx}
\usepackage{textcomp}
\usepackage{xcolor}

\usepackage{pifont}
\usepackage{cleveref}
\usepackage{bbm}
\usepackage[detect-all]{siunitx}
\usepackage{algorithm}
\usepackage[noend]{algpseudocode}
\usepackage{vector}
\usepackage{aircraftshapes}

\usepackage{pgfplots}
\usepgfplotslibrary{groupplots}
\usepgfplotslibrary{fillbetween}

\pgfplotsset{compat=newest}
\pgfplotsset{every axis legend/.append style={legend cell align=left}}
\pgfplotsset{every axis/.append style={
                    title style={font=\small},
                    tick label style={font=\footnotesize}  
                    }}
\pgfplotsset{every axis label/.style={font=\small}}                    
\pgfkeys{/pgf/number format/.cd,std=-2}

\definecolor{pastelMagenta}{HTML}{FF48CF}
\definecolor{pastelPurple}{HTML}{8770FE}
\definecolor{pastelBlue}{HTML}{1BA1EA}
\definecolor{pastelSeaGreen}{HTML}{14B57F}
\definecolor{pastelGreen}{HTML}{3EAA0D}
\definecolor{pastelOrange}{HTML}{C38D09}
\definecolor{pastelRed}{HTML}{F5615C}
\definecolor{blush}{rgb}{0.87, 0.36, 0.51}
\definecolor{darkgreen}{rgb}{0.0392156862745098, 0.35490196078431374, 0.24901960784313726}
\definecolor{cmap1}{rgb}{0.10588235294117647,0.6313725490196078,0.9176470588235294}
\definecolor{cmap2}{rgb}{0.24705882352941178,0.5673202614379085,0.9437908496732026}
\definecolor{cmap3}{rgb}{0.3882352941176471,0.5032679738562091,0.9699346405228758}
\definecolor{cmap4}{rgb}{0.5294117647058824,0.4392156862745098,0.996078431372549}
\definecolor{cmap5}{rgb}{0.673202614379085,0.41960784313725485,0.7843137254901961}
\definecolor{cmap6}{rgb}{0.8169934640522877,0.4,0.5725490196078431}
\definecolor{cmap7}{rgb}{0.9607843137254902,0.3803921568627451,0.3607843137254902}
\definecolor{gray5}{HTML}{696969}
\definecolor{gray4}{HTML}{808080}
\definecolor{gray3}{HTML}{A9A9A9}
\definecolor{gray2}{HTML}{C0C0C0}
\definecolor{gray1}{HTML}{DCDCDC}

\usepackage[backend=bibtex,style=ieee,mincitenames=1,maxcitenames=2,maxbibnames=20,url=false,isbn=false,doi=false]{biblatex}
\def\BibTeX{{\rm B\kern-.05em{\sc i\kern-.025em b}\kern-.08em
    T\kern-.1667em\lower.7ex\hbox{E}\kern-.125emX}}
\begin{document}

\title{Efficient Determination of Safety \\ Requirements for Perception Systems
}

\author{
    \IEEEauthorblockN{Sydney M. Katz, Anthony L. Corso, Esen Yel, and Mykel J. Kochenderfer}
    \IEEEauthorblockA{Department of Aeronautics and Astronautics, Stanford University,
    \{smkatz, acorso, esenyel, mykel\}@stanford.edu}
    \thanks{The NASA University Leadership Initiative (grant \#80NSSC20M0163) provided funds to assist the authors with their research. This research was also supported by the National Science Foundation Graduate Research Fellowship under Grant No. DGE–1656518. Any opinions, findings, and conclusions or recommendations expressed in this material are those of the authors and do not necessarily reflect the views of any NASA entity or the National Science Foundation.}
    }

\maketitle

\begin{abstract}
Perception systems operate as a subcomponent of the general autonomy stack, and perception system designers often need to optimize performance characteristics while maintaining safety with respect to the overall closed-loop system. For this reason, it is useful to distill high-level safety requirements into component-level requirements on the perception system. In this work, we focus on efficiently determining sets of safe perception system performance characteristics given a black-box simulator of the fully-integrated, closed-loop system. We combine the advantages of common black-box estimation techniques such as Gaussian processes and threshold bandits to develop a new estimation method, which we call smoothing bandits. We demonstrate our method on a vision-based aircraft collision avoidance problem and show improvements in terms of both accuracy and efficiency over the Gaussian process and threshold bandit baselines.
\end{abstract}


\section{Introduction}\label{sec:intro}
Perception system designers are often faced with a tradeoff between various performance characteristics such as accuracy and efficiency. Therefore, it is useful to specify a target range of performance requirements for which an automated system will operate safely. For example, consider a camera-based perception system that detects intruder aircraft for the purpose of aircraft collision avoidance \cite{rozantsev2016detecting, james2018learning, opromolla2021visual, ying2022small, ghosh2022airtrack}. This system will have characteristics such as detection probability and field of view. Each possible combination of these performance characteristics will result in a different probability of collision, and it is important to understand the range of these characteristics that will result in an acceptable collision risk.

Our goal in this work is to estimate the set of safe perception system performance characteristics given a probability of failure threshold and a black-box simulator of the fully integrated, closed-loop system. The estimation problem is multi-level. At the highest level, we want to estimate the set of performance characteristics that result in a probability of failure below the threshold. However, producing this estimate requires lower-level estimates of the probability of failure achieved by each possible instantiation of performance characteristics, which we obtain by running simulations using the black-box simulator. 

A na\"ive approach to determining this set is to discretize the space of possible performance characteristics into a grid and run a Monte Carlo analysis at each grid point to check safety requirements. However, Monte Carlo evaluations are often computationally expensive on their own, so performing a Monte Carlo analysis at every point in a grid quickly becomes intractable, especially in high-dimensional problems. In this work, we develop a more efficient way to enumerate the safe set that is inspired by common black-box estimation techniques including Gaussian processes and multi-armed bandits. We develop a technique called smoothing bandits, which combines elements from both techniques to address the key properties of the estimation problem. We apply our method to estimate performance requirements for a realistic vision-based aircraft collision avoidance problem and show that it outperforms the Gaussian process and multi-armed bandit baselines in terms of accuracy and efficiency.

\section{Related work}\label{sec:related_work}
Our safe set estimation approach is inspired by techniques used for level set estimation (LSE) \cite{bryan2005active, gotovos2013active, bogunovic2016truncated, zanette2018robust, iwazaki2020bayesian, inatsu2021active, neiswanger2021bayesian} and threshold bandits \cite{chen2014combinatorial, locatelli2016optimal, abernethy2016threshold, jamieson2018bandit}.

\subsection{Level set estimation}\label{subsec:lse}
LSE aims to determine the set of inputs for which the output of a function exceeds a given threshold \cite{willett2007minimax}. In the context of this work, this formulation is analogous to determining the set of performance characteristics that result in a desired level of safety. Because evaluating the safety of a given set of performance characteristics requires a Monte Carlo analysis, we can treat the problem as a black-box estimation problem with expensive function evaluations. In this regard, LSE is related to the field of Bayesian optimization. Bayesian optimization aims to optimize an unknown function by fitting a probabilistic surrogate model using Gaussian process regression and selecting where to sample next based on an acquisition function \cite{shahriari2015taking, frazier2018tutorial}. While the focus of Bayesian optimization is to find global optima of unknown functions, the focus of LSE is to find all inputs to an unknown function that result in outputs that exceed a given threshold. Multiple recent works have proposed techniques to modify traditional Bayesian optimization techniques and acquisition functions to perform well in the LSE context \cite{bryan2005active, gotovos2013active, bogunovic2016truncated, zanette2018robust, iwazaki2020bayesian, inatsu2021active, neiswanger2021bayesian}.


\subsection{Threshold bandits}\label{subsec:threshold_bandits}
In the same way that LSE is related to Bayesian optimization, threshold bandits are related to multi-armed bandits. The multi-armed bandit problem models situations in which an agent must select from a series of arms to pull, each with an unknown reward distribution \cite{alg4dm}. The agent's goal is to maximize expected reward given a fixed number of pulls, requiring a balance between exploration and exploitation. While the traditional multi-armed bandit setting aims to find the arm with the best expected reward, the threshold bandit setting aims to find all arms with an expected reward that exceeds a given threshold. This formulation has been used in clinical trials, where the goal is to select a set of promising drugs to move to the next phase of testing \cite{katehakis1986computing, jamieson2018bandit}. In the context of this work, we can model each simulation in a Monte Carlo analysis as a pull of an arm in a multi-armed bandit problem and use threshold bandit techniques to efficiently estimate safe performance requirements. Threshold bandits are a subproblem of the combinatorial pure exploration problem studied by \citeauthor{chen2014combinatorial} \cite{chen2014combinatorial}, and other variations of the problem have recently been studied~\cite{locatelli2016optimal, abernethy2016threshold, jamieson2018bandit, mason2020finding}.

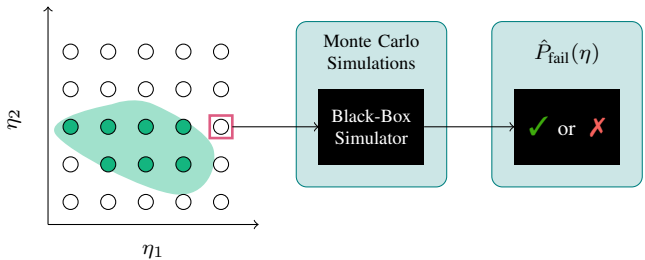
\begin{figure}[t]
    \centering
    \begin{tikzpicture}

    \fill[pastelSeaGreen!40]  plot[smooth, tension=.7] coordinates {(-1.95,-0.2) (-1.4,-0.2) (-0.6,-0.8) (-1.1,-1.4) (-2.7,-0.6) (-1.95,-0.2)};

    \draw (-0.5, 0.5) circle (0.1);
    \draw (-0.5, 0.0) circle (0.1);
    \draw (-0.5, -0.5) circle (0.1);
    \draw (-0.5, -1.0) circle (0.1);
    \draw (-0.5, -1.5) circle (0.1);

    \draw (-1.0, 0.5) circle (0.1);
    \draw (-1.0, 0.0) circle (0.1);
    \draw[fill=pastelSeaGreen] (-1.0, -0.5) circle (0.1);
    \draw[fill=pastelSeaGreen] (-1.0, -1.0) circle (0.1);
    \draw (-1.0, -1.5) circle (0.1);

    \draw (-1.5, 0.5) circle (0.1);
    \draw (-1.5, 0.0) circle (0.1);
    \draw[fill=pastelSeaGreen] (-1.5, -0.5) circle (0.1);
    \draw[fill=pastelSeaGreen] (-1.5, -1.0) circle (0.1);
    \draw (-1.5, -1.5) circle (0.1);

    \draw (-2.0, 0.5) circle (0.1);
    \draw (-2.0, 0.0) circle (0.1);
    \draw[fill=pastelSeaGreen] (-2.0, -0.5) circle (0.1);
    \draw[fill=pastelSeaGreen] (-2.0, -1.0) circle (0.1);
    \draw (-2.0, -1.5) circle (0.1);

    \draw (-2.5, 0.5) circle (0.1);
    \draw (-2.5, 0.0) circle (0.1);
    \draw[fill=pastelSeaGreen] (-2.5, -0.5) circle (0.1);
    \draw (-2.5, -1.0) circle (0.1);
    \draw (-2.5, -1.5) circle (0.1);


    \draw[->] (-2.8, -1.8) -- (0.0, -1.8);
    \node[anchor=north] at (-1.4, -1.95) {\footnotesize $\eta_1$};
    \draw[->] (-2.8, -1.8) -- (-2.8, 1.1);
    \node[rotate=90] at (-3.25, -0.4) {\footnotesize $\eta_2$};

    \fill[rounded corners, teal!20, draw=teal] (0.5, -1.3) rectangle (2.5, 0.9);
    \node at (1.5, 0.65) {\scriptsize Monte Carlo};
    \node at (1.5, 0.35) {\scriptsize Simulations};

    \fill[rounded corners, teal!20, draw=teal] (3.1, -1.3) rectangle (5.1, 0.9);
    \node at (4.1, 0.5) {\footnotesize $\hat P_\text{fail}(\eta)$};

    \fill[black] (0.8, -1.0) rectangle (2.2, 0.0);
    \node[white] at (1.5, -0.35) {\scriptsize Black-Box};
    \node[white] at (1.5, -0.65) {\scriptsize Simulator};

    \fill[black] (3.4, -1.0) rectangle (4.8, 0.0);
    \node at (4.1, -0.5) {\textcolor{pastelGreen}{\ding{51}} \footnotesize\textcolor{white}{or} \ \normalsize\textcolor{pastelRed}{\ding{55}}};

    \draw[blush, line width=1.0] (-0.65, -0.65) rectangle (-0.35, -0.35);
    \draw[->] (-0.35, -0.5) -- (0.8, -0.5);

    \draw[->] (2.2, -0.5) -- (3.4, -0.5);
    
\end{tikzpicture}
    \caption{Overview of the safety requirement estimation problem. \label{fig:req_overview}}
\end{figure}

\section{Problem Formulation}\label{sec:prob_form}
We consider the setting shown in \cref{fig:req_overview}. An automated system has performance characteristics represented by $\eta \in \mathbb R^d$ where $d$ is the number of controllable performance characteristics. We assume access to a black-box simulator $f_\text{bb}(\eta)$ that takes in a particular instantiation of $\eta$ and simulates an episode of the closed-loop behavior of a stochastic dynamical system. For a single simulation episode, the output of this simulator can be compared to a given safety specification $\psi$ to determine whether a failure occurred. Therefore, the probability of failure for a particular value of $\eta$ is represented as
\begin{equation} \label{eq:req_pfail}
    P_\text{fail}(\eta) = \mathbb E [\mathbbm 1 \{f_\text{bb}(\eta) \notin \psi \} ]
\end{equation}
where the expectation is governed by the stochastic nature of the closed-loop dynamical system. Our goal is then to determine the set of performance characteristics $I \subseteq \mathbb R^d$ for which the probability of failure is below a given threshold as follows
\begin{equation} \label{eq:safeset}
    I = \{\eta \mid P_\text{fail}(\eta) < \gamma \}
\end{equation}
where $P_\text{fail}(\eta)$ represents the probability of failure given performance characteristics $\eta$ and $\gamma$ is a pre-determined safety threshold.

Due to the black-box nature of the problem, calculating an exact value for the probability of failure is not possible. However, \cref{eq:req_pfail} can be approximated via Monte Carlo analysis using the black-box simulator as follows
\begin{equation} \label{eq:approx_pfail}
    \hat P_\text{fail}(\eta) \approx \frac{1}{N} \sum_{i=1}^N \mathbbm 1 \{f_\text{bb}(\eta) \notin \psi \}
\end{equation}
where $N$ is the number of simulation episodes. Therefore, a simple way to estimate $I$ is to discretize the space of possible parameter values into a set of grid points $\Omega$. For each point $\eta \in \Omega$, we can solve for $\hat P_\text{fail}(\eta)$ using \cref{eq:approx_pfail} and compare the result to $\gamma$ to determine whether to include the point in the safe set $I$. For large values of $N$, this technique will provide accurate estimates of the safe set; however, simulating a large number of episodes even for a single point $\eta \in \Omega$ is computationally expensive and doing so for all points in $\Omega$ may be intractable. In this work, we develop a more efficient approach for estimating $I$.

\subsection{Pendulum example} 
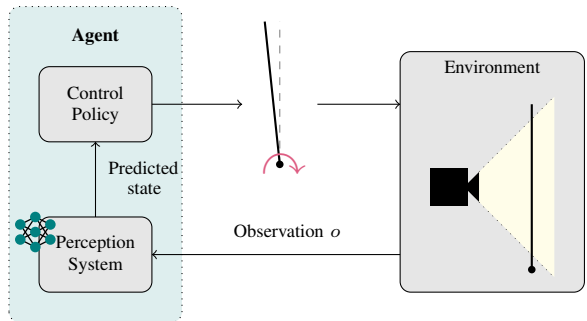
\begin{figure}[t]
    \centering
    \begin{tikzpicture}

    \filldraw[rounded corners, fill=teal!13, dotted, draw=black] (-0.4, -0.4) rectangle (1.9, 3.8);
    \node at (0.75, 3.4) {\scriptsize \textbf{Agent}};

    \filldraw[fill=gray!20, draw=black, rounded corners] (0.0, 0.0) rectangle (1.5, 1.0);
    \node at (0.75, 0.65) {\scriptsize Perception};
    \node at (0.75, 0.35) {\scriptsize System};

    \filldraw[fill=gray!20, draw=black, rounded corners] (0.0, 2.0) rectangle (1.5, 3.0);
    \node at (0.75, 2.65) {\scriptsize Control};
    \node at (0.75, 2.35) {\scriptsize Policy};

    \draw[->] (0.75, 1.0) -- (0.75, 2.0);
    \node[anchor=west] at (0.8, 1.65) {\scriptsize Predicted};
    \node at (1.4, 1.38) {\scriptsize state};

    \filldraw[fill=gray!20, draw=black, rounded corners] (4.8, 0.0) rectangle (7.25, 3.2);
    \node at (6.025, 3.0) {\scriptsize Environment};

    \fill (5.2, 1.15) rectangle (5.7, 1.65);
    \fill (5.7, 1.45) -- (5.85, 1.6) -- (5.85, 1.2) -- (5.7, 1.35);
    \draw[dotted] (5.85, 1.6) -- (6.85, 2.6);
    \draw[dotted] (5.85, 1.2) -- (6.85, 0.2);
    \fill[fill = yellow!10] (5.85, 1.6) -- (6.85, 2.6) -- (6.85, 0.2) -- (5.85, 1.2) -- cycle;

    \fill (6.55, 0.3) circle (0.05);
    \draw[thick] (6.55, 0.3) -- (6.55, 2.5);



    \draw[->] (1.5, 2.5) -- (2.7, 2.5);

    \fill (3.2, 1.7) circle (0.05);
    \draw[thick] (3.2, 1.7) -- (3.0, 3.6);
    \draw[dashed, opacity=0.5] (3.2, 1.7) -- (3.2, 3.6);
    \node at (3.25, 1.7) {\color{blush} \LARGE $\curvearrowright$};

    \draw[->] (3.7, 2.5) -- (4.8, 2.5);

    \draw[->] (4.8, 0.5) -- (1.5, 0.5);
    \node[anchor=south] at (3.3, 0.6) {\scriptsize Observation $o$};

    \draw[line width=0.5] (-0.25, 0.9) -- (-0.05, 1.0);
    \draw[line width=0.5] (-0.25, 0.9) -- (-0.05, 0.8);
    \draw[line width=0.5] (-0.25, 0.9) -- (-0.05, 0.6);

    \draw[line width=0.5] (-0.25, 0.7) -- (-0.05, 1.0);
    \draw[line width=0.5] (-0.25, 0.7) -- (-0.05, 0.8);
    \draw[line width=0.5] (-0.25, 0.7) -- (-0.05, 0.6);

    \draw[line width=0.5] (-0.05, 1.0) -- (0.15, 0.9);
    \draw[line width=0.5] (-0.05, 1.0) -- (0.15, 0.7);

    \draw[line width=0.5] (-0.05, 0.8) -- (0.15, 0.9);
    \draw[line width=0.5] (-0.05, 0.8) -- (0.15, 0.7);

    \draw[line width=0.5] (-0.05, 0.6) -- (0.15, 0.9);
    \draw[line width=0.5] (-0.05, 0.6) -- (0.15, 0.7);
    
    \fill[teal] (-0.25, 0.7) circle (0.07);
    \fill[teal] (-0.25, 0.9) circle (0.07);

    \fill[teal] (-0.05, 0.6) circle (0.07);
    \fill[teal] (-0.05, 0.8) circle (0.07);
    \fill[teal] (-0.05, 1.0) circle (0.07);

    \fill[teal] (0.15, 0.7) circle (0.07);
    \fill[teal] (0.15, 0.9) circle (0.07);

\end{tikzpicture}
    \caption{Overview of the inverted pendulum example problem. \label{fig:pend_overview}}
\end{figure}
\begin{figure}[t]
    \centering
    \input{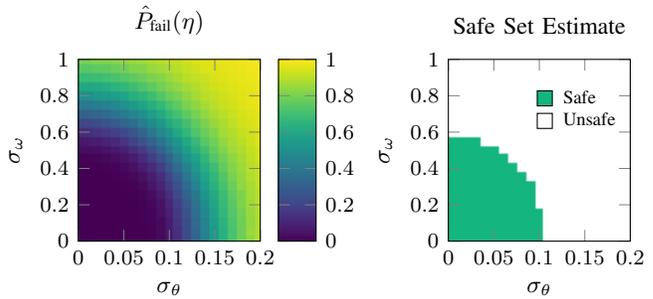}
    \caption{Ground truth results for the pendulum. \label{fig:pend_gt}}
\end{figure}
We use a vision-based inverted pendulum as a running example for our approach. \Cref{fig:pend_overview} provides an overview of this problem. We assume a camera produces image observations of the pendulum's current state. We represent the state of the pendulum as $s=[\theta, \omega]$, where $\theta$ is its angle from the vertical and $\omega$ is its angular velocity. A perception system (such as a neural network) produces an estimate of these state variables, which gets passed to a controller that selects a torque to keep the pendulum upright. We define the safety property to require that $|\theta|< \pi / 4$ at all time steps. We assume that the perception errors for each state variable are independent and normally distributed with zero mean and standard deviation $\sigma_\theta$ and $\sigma_\omega$. We select these values as our performance characteristics such that $\eta = [\sigma_\theta, \sigma_\omega]$. 

We create a black-box simulator that takes in a value for $\eta$ and simulates an episode starting from a random initial state. After each episode, the simulator checks whether the trajectory satisfies the safety property and outputs the result. The low-dimensionality of the problem allows us to apply a na\"ive Monte Carlo approach to estimate the probability of failure and the safe set given a threshold of $\gamma=0.1$. We discretize the space of possible values for $\eta$ into a $21 \times 21$ grid and simulate \num{10000} episodes at each point. \Cref{fig:pend_gt} shows the results, which we will use as ground truth to evaluate the more efficient methods we develop in the rest of this work.

\section{Black-box estimation of safe sets}
\begin{algorithm}[b]
    \caption{Black-Box Estimation of Safe Sets \label{alg:safe_set_est}}
    \begin{algorithmic}[1]
        \Function{SafeSetEstimation}{$f_\text{bb}$, $\Omega$, $\gamma$, $\delta$, $P_0$}
            \State $P(P_\text{fail}) \gets P_0$
            \State Sample initial point $\eta^\prime \in \Omega$
            \Loop
               \State result $\gets$ \Call{Simulate}{$f_\text{bb}$, $\eta^\prime$}
               \State $P(P_\text{fail}) \gets$ \Call{Update}{$P(P_\text{fail})$, result}
               \State $\eta^\prime \gets$ \Call{Acquisition}{$P(P_\text{fail})$, $\Omega$}
            \EndLoop
            \State \textbf{return} \Call{Extract Safe Set}{$P(P_\text{fail})$, $\Omega$, $\gamma$, $\delta$}
        \EndFunction
    \end{algorithmic}
\end{algorithm}
\Cref{alg:safe_set_est} outlines a general approach for black-box estimation of safe sets. The algorithm takes in the black-box simulator $f_\text{bb}$, the set of candidate points $\Omega$, the safety threshold $\gamma$, a confidence threshold $\delta$, and a prior distribution over $P_\text{fail}(\eta)$ denoted as $P_0$. At each iteration, we simulate a candidate point using the black-box simulator and obtain the results. We then use these results to update the distribution over $P_\text{fail}(\eta)$, and call an acquisition function on this new distribution to determine which point to evaluate next. 

After iterating through a fixed budget of simulation episodes, we robustly estimate the safe set $I$ given the current distribution over $P_\text{fail}(\eta)$ and a desired confidence $\delta$. Specifically, we require that
\begin{equation}\label{eq:robust_extract}
    P(P_\text{fail}(\eta) < \gamma) > \delta
\end{equation}
holds for inclusion in the safe set. In effect, $\delta$ controls the precision of the estimation technique, which is related to the number of false positives. Our goal is then to maximize the number of points in $\Omega$ that we can classify as safe with as few queries to the black-box simulator as possible. To summarize, each approach to black-box estimation of safe sets must define the following three components: a distribution over $P_\text{fail}(\eta)$, a safe set extraction method given $P_\text{fail}(\eta)$, and an acquisition function. The following sections will outline these components for two baseline methods (level set estimation using Gaussian processes and threshold bandits) and our proposed method, which combines the advantages of the baselines.

\section{Baseline methods}\label{sec:baselines}
Our approach is inspired by two baseline techniques that are commonly used for black-box estimation tasks: Gaussian processes and threshold bandits. This section outlines how these techniques can be used to solve the desired estimation problem.

\subsection{Level set estimation using Gaussian processes}
In this approach, we assume that each iteration of the loop in \cref{alg:safe_set_est} runs a fixed number of episodes in the black-box simulator to estimate $P_\text{fail}(\eta)$ using \cref{eq:approx_pfail}. We model the distribution over $P_\text{fail}(\eta)$ as a Gaussian process and use it to robustly estimate the safe set and select evaluation points. The rest of this section outlines the components of \cref{alg:safe_set_est} for the Gaussian process baseline.

\paragraph{Model} Gaussian processes estimate black-box functions by maintaining a distribution over possible functions~\cite{alg4opt}. In the context of this work, we can use a Gaussian process to model a distribution over the function $P_\text{fail}(\eta)$. Consider a set of $n$ points $\vect \eta = [\eta^{(1)}, \ldots, \eta^{(n)}]$ that we have evaluated in our black-box simulator to obtain probability of failure estimates $\vect P = [\hat P_\text{fail}(\eta^{(1)}), \ldots, \hat P_\text{fail}(\eta^{(n)})]$. Suppose we have a set of $m$ new points $\vect \eta^\prime = [\eta^{\prime(1)}, \ldots, \eta^{\prime(m)}]$ for which we would like to estimate $\vect P^\prime = [\hat P_\text{fail}(\eta^{\prime(1)}), \ldots, \hat P_\text{fail}(\eta^{\prime(m)})]$. Given a mean function $m(\eta)$ and a kernel function $k(\eta, \eta^\prime)$, Gaussian processes model the joint distribution between $\vect P$ and $\vect P^\prime$ as a multivariate Gaussian distribution. We assume that the estimates of the probability of failure are noisy evaluations with $\hat P_\text{fail}(\eta) = f_\text{bb}(\eta) + w$ where $w \sim \mathcal N(0, \nu)$. For this work, we select $\nu$ using the coefficient of variation of the estimator described in \cref{eq:approx_pfail}.

Given this model, we can estimate a distribution over the probability of failure at points $\vect \eta^\prime$ by computing the mean and covariance of the conditional distribution $P(\vect P^\prime \mid \vect P, \nu)$, which will also be Gaussian. The mean function can be selected based on prior knowledge and the kernel function is selected to control the smoothness of the estimated functions. In this work, we select $m(\eta) = 0$ and use a weighted squared exponential kernel, a common kernel for Gaussian processes defined as
\begin{equation}\label{eq:wsqe_kernel}
    k(\eta, \eta^\prime) = \exp \left(\frac{(\eta - \eta^\prime)^\top W (\eta - \eta^\prime)}{2 \ell^2}\right)
\end{equation}
where $W$ is a weighting matrix and $\ell$ is a length parameter that controls smoothness. Intuitively, $k(\eta, \eta^\prime)$ outputs a higher value as $\eta$ and $\eta^\prime$ get closer in Euclidean distance. A higher length parameter indicates a smoother function and therefore assumes higher correlation between nearby datapoints.

\paragraph{Safe Set Extraction} Using the Gaussian process model, we can decide whether to include a point in the safe set based on its posterior probability distribution and desired confidence $\delta$. Specifically, let $\mu_{GP}(\eta)$ and $\sigma_{GP}(\eta)$ represent the posterior mean and standard deviation respectively for a point $\eta$ given the Gaussian process model. We want to include points in the safe set according to \cref{eq:robust_extract}, which can be written in terms of the posterior mean and variance as follows
\begin{equation} \label{eq:gp_decidesafe}
    \mu_{GP}(\eta) + \beta \sigma_{GP}(\eta) \leq \gamma
\end{equation}
where $\beta$ is the multiplier for the standard deviation required to capture a $\delta$-fraction of the probability mass  \cite{zanette2018robust}.

\paragraph{Acquisition Function} Bayesian optimization uses acquisition functions to select future evaluation points. An acquisition function takes in the current distribution over the function and automatically selects a new point for evaluation. In traditional Bayesian optimization where the objective is to find the global optimum of an unknown function, acquisition functions typically balance between selecting points that are likely to be near optimal and selecting points with high uncertainty. Recent work has focused on adapting these acquisition functions to scenarios where the primary objective is to determine the level set of an unknown function \cite{bryan2005active, gotovos2013active, bogunovic2016truncated, zanette2018robust, iwazaki2020bayesian, inatsu2021active, neiswanger2021bayesian}. As a baseline in this work, we will use the maximum improvement in level-set estimation (MILE) acquisition function developed by \citeauthor{zanette2018robust} \cite{zanette2018robust}, which selects the point that results in the highest expected improvement in the safe set size. 

\paragraph{Pendulum Example}
\begin{figure}[t]
    \centering
    \input{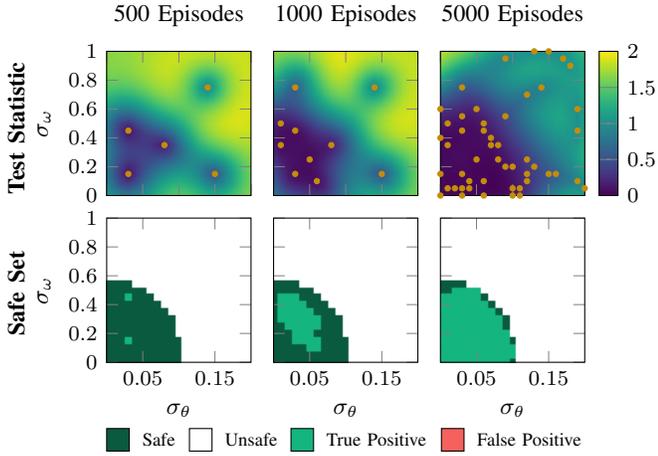}
    \vspace{-2mm}
    \caption{Pendulum Gaussian process MILE acquisition. Orange points represent evaluated points. \label{fig:pend_gp_mile}}
\end{figure}
\Cref{fig:pend_gp_mile} shows the progression of the Gaussian process model and safe set estimate using the MILE acquisition function. The top row shows the test statistic in \cref{eq:gp_decidesafe} for $\delta=0.95$, and the bottom row shows the corresponding estimated and ground truth safe set. The test statistic is high in regions with few evaluations due to high uncertainty. Most of the ground truth safe set has been properly identified after only \num{5000} episodes, which is less than \SI{1}{\percent} the number of episodes used in the na\"ive Monte Carlo approach. Moreover, the algorithm has avoided false positives due to its robust selection of safe points.

\subsection{Threshold Bandits}\label{sec:threshold_bandits}
In this approach, we assume that each iteration of the loop in \cref{alg:safe_set_est} runs a single simulation episode in the black-box simulator, corresponding to a pull of an arm in a multi-armed bandit problem. We model $P_\text{fail}(\eta)$ as a set of unknown reward distributions for arms $\eta \in \Omega$ and use this model to robustly estimate the safe set and select evaluation points. The rest of this section outlines the components of \cref{alg:safe_set_est} for the threshold bandit baseline.

\paragraph{Model} The multi-armed bandit problem refers to scenarios in which an agent is given a fixed number of pulls and must select an arm to pull at each time step from a set of arms with unknown reward distributions \cite{alg4dm}. In this work, we will focus on Bernoulli bandits, where pulling arm $\eta$ results in a success with probability $1 - P_\text{fail}(\eta)$ and a failure otherwise. We will denote the probability of success for each arm as $\theta_\eta = 1 - P_\text{fail}(\eta)$. We can model the distribution over $\theta_\eta$ (and implicitly the distribution over $P_\text{fail}(\eta)$) as a beta distribution. If we assume a uniform prior of $\text{Beta}(1, 1)$, the posterior distribution over $\theta_\eta$ can be calculated according to
\begin{equation}\label{eq:bandit_post}
    \theta_\eta \sim \text{Beta}(1 + p_\eta, 1 + q_\eta)
\end{equation}
where $p_\eta$ represents the number of black-box simulation episodes with performance characteristics $\eta$ that resulted in a satisfaction of the safety property and $q_\eta$ represents the number of episodes that resulted in a safety violation. One limitation of the bandit model compared to the Gaussian process model is that the bandit model does not define $P_\text{fail}(\eta)$ over the entire continuous space $\eta \in \mathbb R^d$ but rather at the discrete points $\eta \in \Omega$. To account for points not in $\Omega$, we use nearest neighbor interpolation.

\begin{figure}[t]
    \centering
    \input{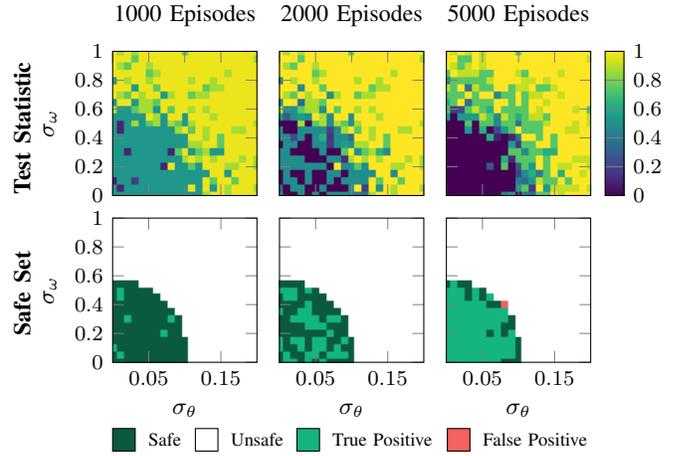}
    \caption{Pendulum bandit DKWUCB acquisition. \label{fig:pend_bandit_acq}}
\end{figure}

\begin{figure*}[t]
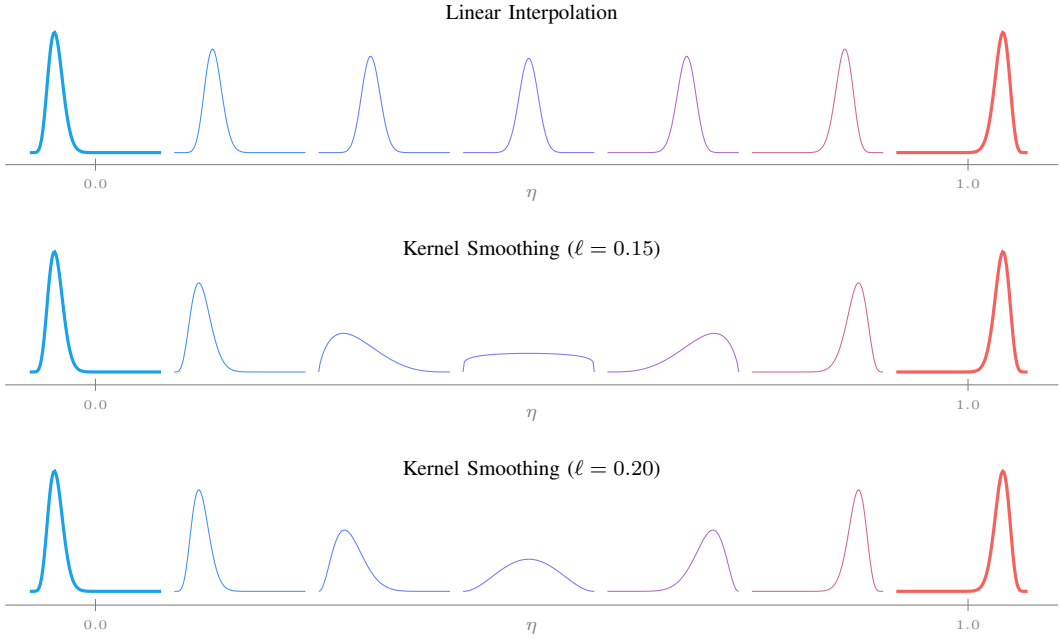

    \centering
    \begin{tikzpicture}[]


\begin{groupplot}[group style={horizontal sep = 3.5cm, vertical sep = 1.0cm, group size=1 by 3}]

\nextgroupplot [
  height = {3.5cm},
  hide axis,
  width = {17.5cm},
]

\input{figures/subplots/interp_points}

\nextgroupplot [
  height = {3.5cm},
  hide axis,
  width = {17.5cm},
]

\input{figures/subplots/kernel_points_1}

\nextgroupplot [
  height = {3.5cm},
  hide axis,
  width = {17.5cm},
]

\input{figures/subplots/kernel_points_2}

\end{groupplot}

\draw[gray] (1.0, 0.0) -- (15.0, 0.0);
\node[gray] at (8.0, -0.4) {\scriptsize $\eta$};
\draw[gray] (2.2, -0.08) -- (2.2, 0.08);
\draw[gray] (13.8, -0.08) -- (13.8, 0.08);
\node[gray, anchor=north] at (2.2, -0.08) {\tiny $0.0$};
\node[gray, anchor=north] at (13.8, -0.08) {\tiny $1.0$};

\draw[gray] (1.0, -2.93) -- (15.0, -2.93);
\node[gray] at (8.0, -3.33) {\scriptsize $\eta$};
\draw[gray] (2.2, -3.01) -- (2.2, -2.85);
\draw[gray] (13.8, -3.01) -- (13.8, -2.85);
\node[gray, anchor=north] at (2.2, -3.01) {\tiny $0.0$};
\node[gray, anchor=north] at (13.8, -3.01) {\tiny $1.0$};

\draw[gray] (1.0, -5.86) -- (15.0, -5.86);
\node[gray] at (8.0, -6.16) {\scriptsize $\eta$};
\draw[gray] (2.2, -5.94) -- (2.2, -5.78);
\draw[gray] (13.8, -5.94) -- (13.8, -5.78);
\node[gray, anchor=north] at (2.2, -5.94) {\tiny $0.0$};
\node[gray, anchor=north] at (13.8, -5.94) {\tiny $1.0$};

\node at (8.0, 2.0) {\footnotesize Linear Interpolation};
\node at (8.0, -1.13) {\footnotesize Kernel Smoothing ($\ell=0.15$)};
\node at (8.0, -4.06) {\footnotesize Kernel Smoothing ($\ell=0.20$)};


\end{tikzpicture}
    \caption{Comparison of local approximation methods. \label{fig:kernel_demo}}
\end{figure*}

\paragraph{Safe Set Extraction} We can determine whether to include points in the safe set based on their estimated posterior probability distribution. To decide whether to include a given point $\eta$, we rewrite the requirement in \cref{eq:robust_extract} for the bandit case as
\begin{equation}\label{eq:bandit_decide_safe}
    F^{-1}_\eta(\delta) \leq \gamma
\end{equation}
where $F^{-1}_\eta$ is the inverse cumulative distribution function for the beta distribution over $P_\text{fail}(\eta)$. Intuitively, we require that a $\delta$-fraction of the probability mass of the distribution over the probability of failure lies below the required threshold. 

\paragraph{Acquisition Function} Similar to the acquisition functions used for Gaussian processes, strategies for arm selection in multi-armed bandits balance between exploration and exploitation. Common ad hoc strategies include $\epsilon$-greedy, softmax exploration, UCB1 exploration, and Thompson sampling~\cite{alg4dm}. Recent work has focused on adapting these strategies for threshold bandits \cite{chen2014combinatorial, locatelli2016optimal, abernethy2016threshold, jamieson2018bandit}. In this work, we use a modified version of the DKWUCB algorithm developed by \citeauthor{abernethy2016threshold} \cite{abernethy2016threshold}, which was inspired by the UCB1 algorithm. At each iteration, we select the arm that maximizes
\begin{equation}\label{eq:dkwucb}
    F_\eta(\gamma) + \begin{cases}
    \sqrt{\frac{\log(2/c)}{2N_\eta}}, & N_\eta > 0 \text{ and } \eta \notin \hat I \\
    -\infty, & \eta \in \hat I \\
    1, & \text{otherwise} \\
    \end{cases}
\end{equation}
where $F_\eta$ is the cumulative distribution function for the distribution over $P_\text{fail}(\eta)$, $N_\eta = p_\eta + q_\eta$ is the number of simulation episodes for arm $\eta$, $\hat I$ is the estimated safe set based on the current distribution over $P_\text{fail}(\eta)$, and $c > 0$ is a tunable exploration constant. Intuitively, the acquisition function balances between selecting arms that are likely to be safe and arms that are under-explored. As soon as we have enough confidence to include an arm in the safe set, \cref{eq:dkwucb} will evaluate to $-\infty$, and we will no longer select it.

\paragraph{Pendulum Example} 
\Cref{fig:pend_bandit_acq} shows the progression of the DKWUCB algorithm for the pendulum problem using exploration constant $c=1$. The top row shows the test statistic from \cref{eq:bandit_decide_safe} and the bottom row shows the safe set estimate. After \num{5000} episodes, the algorithm has identified most points in the safe set. There is one false positive near the edge of the safe set, which is expected given the precision implied by setting $\delta=0.95$. The safe set estimates are more disjoint for the bandit formulation than for the Gaussian process formulation. This result is due to the lack of spatial correlation in the bandit model.

\subsection{Baseline Comparison} The Gaussian process and threshold bandit formulations each have distinct advantages. Gaussian processes, for example, use a kernel function to account for spatial correlation between nearby function evaluations. Since similar perception performance characteristics are likely to exhibit similar safety properties, modeling this correlation can greatly increase the efficiency of the estimation technique. One drawback of Gaussian process approaches is that they cannot easily incorporate the multi-level nature of the estimation problem in the way that threshold bandits can. 

While the Gaussian process formulation assumes each function evaluation involves a fixed number of simulations, the threshold bandit formulation assumes each pull of an arm is a single simulation episode. Therefore, the threshold bandit formulation has the flexibility to allocate more simulations to points that are likely to be safe or have high uncertainty. On the other hand, the threshold bandit formulation does not take into account spatial correlation between the points. In \cref{sec:smoothing_bandits}, we outline an approach that combines elements from both baselines to simultaneously address both spatial correlation and the multi-level nature of the estimation problem.

\section{Smoothing bandits}\label{sec:smoothing_bandits}
In this section, we develop a new model that combines the advantages of the baselines, which we will refer to as smoothing bandits. As in \cref{sec:threshold_bandits}, we model the problem as a multi-armed bandit problem and assume we run one simulation episode in the black-box simulator at each iteration. To account for spatial correlation, we apply kernel smoothing to the observed data.

\subsection{Smoothing Bandit Model}
As in \cref{sec:threshold_bandits}, we use a Bernoulli bandit model with arms $\eta \in \Omega$ and model the distribution over $P_\text{fail}(\eta)$ for a given arm $\eta$ as a beta distribution. We let $p_\eta$ and $q_\eta$ represent the number of observed success and failure episodes at point $\eta$ respectively and determine the posterior probability distribution over $P_\text{fail}(\eta)$ using \cref{eq:bandit_post}. This model assumes the arms are completely independent; however, it is likely that arms near each other in terms of performance characteristics will have similar probabilities of failure. To account for this correlation, we use local approximation techniques. For example, consider the scenario shown in \cref{fig:kernel_demo}. In this scenario, the observed counts at $\eta=0.0$ have resulted in the blue distribution over the probability of failure (left), and the observed counts at $\eta=1.0$ have resulted in the red distribution (right). 

We can fill in the probability of failure distributions at the intermediate points by applying linear interpolation to the observed counts, but this technique results in some undesirable behavior. As shown in the top row of \cref{fig:kernel_demo}, the intermediate distributions all have high confidence in a small range of failure probabilities. However, as we move away from observed data points, the distributions should tend toward greater entropy to reflect higher uncertainty in the predictions. To achieve this property, we instead apply another form of local approximation to the observed data called kernel smoothing.

Similar to the Gaussian process model, kernel smoothing relies on a kernel function $k(\eta, \eta^\prime)$ and corresponding kernel matrix $\mathbf K (\vect \eta, \vect \eta^\prime)$ defined as
\begin{equation}\label{eq:kernel_matrix}
    \mathbf K (\vect \eta, \vect \eta^\prime) = \begin{bmatrix}
        k(\eta^{(1)}, \eta^{\prime(1)}) & \ldots & k(\eta^{(1)}, \eta^{\prime(m)}) \\ 
        \vdots & \ddots & \vdots \\ 
        k(\eta^{(n)}, \eta^{\prime(1)}) & \ldots & k(\eta^{(n)}, \eta^{\prime(m)})
    \end{bmatrix}
\end{equation}
Consider a set of $n$ points $\vect \eta = [\eta^{(1)}, \ldots, \eta^{(n)}]$ for which we have success counts $\vect p = [p_{\eta^{(1)}}, \ldots, p_{\eta^{(n)}}]$ and failure counts $\vect q = [q_{\eta^{(1)}}, \ldots, q_{\eta^{(n)}}]$. Suppose we have a set of $m$ new points $\vect \eta^\prime = [\eta^{\prime(1)}, \ldots, \eta^{\prime(m)}]$ for which we would like to estimate the failure probability distribution. We can apply kernel smoothing as follows
\begin{equation}
    \begin{split}
        \vect p^\prime & = \mathbf K (\vect \eta, \vect \eta^\prime) \vect p \\
        \vect q^\prime & = \mathbf K (\vect \eta, \vect \eta^\prime) \vect q
    \end{split}
\end{equation}
where $\vect p^\prime$ and $\vect q^\prime$ are the estimated counts, which we can use to define the failure probability distributions for the points in $\vect \eta^\prime$. \Cref{fig:kernel_demo} shows the results of applying kernel smoothing using two different length parameters for the weighted squared exponential kernel defined in \cref{eq:wsqe_kernel}. As we move away from the observed points, the distributions decrease in certainty and approach a uniform distribution.

For the safe set estimation problem, we apply the kernel to the observed counts to estimate a set of smoothed counts for all points in $\Omega$. Let $\vect \eta$ represent all points in $\Omega$, and let $\vect p$ and $\vect q$ represent vectors of the corresponding observed successes and failures. We define the estimated counts as follows
\begin{equation}\label{eq:est_counts}
    \begin{split}
        \mathbf{\hat p} & = \mathbf K (\vect \eta, \vect \eta) \vect p \\
        \mathbf{\hat q} & = \mathbf K (\vect \eta, \vect \eta) \vect q
    \end{split}
\end{equation}
We note that the effective number of observations at a given point $\eta$ will increase 
using this model. In a way, we are allowing arms to share their neighbors' results. Observing \num{10} successes at a neighboring point also provides evidence at the current point and will increase the success count at the current point by some number less than \num{10}.

\paragraph{Pendulum Example} 
\begin{figure}[t]
    \centering
    \input{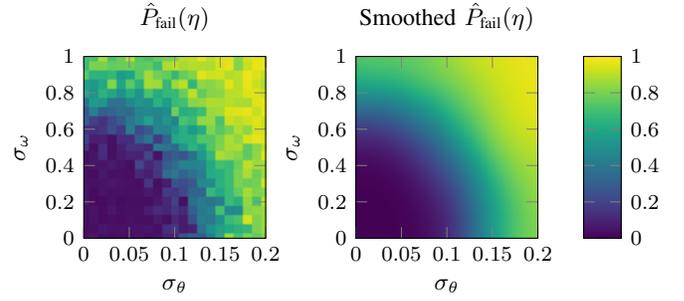}
    \vspace{-3mm}
    \caption{Pendulum smoothing bandit model. \label{fig:pend_smooth_model}}
    \vspace{-3mm}
\end{figure}
\Cref{fig:pend_smooth_model} shows the expectation of the distribution over $P_\text{fail}(\eta)$ after simulating \num{10000} pulls (episodes) distributed randomly among the points in $\Omega$. We show the result for both the raw observations as well as the smoothed observations using a weighted squared exponential kernel with length parameter $\ell=0.02$. The smoothed results closely match the true failure probabilities.

\subsection{Smoothing Bandit Safe Set Extraction and Acquisition} 
We can adapt the methods used in \cref{sec:threshold_bandits} to the smoothing bandit setting with only small modifications. We determine whether to include a given point $\eta$ using \cref{eq:bandit_decide_safe}, but the distribution over $P_\text{fail}(\eta)$ is determined by the smoothed counts rather than the observed counts. We also use the DKWUCB acquisition function defined in \cref{eq:dkwucb} with $F_\eta$ and $N_\eta$ defined by the smoothed counts.

\paragraph{Pendulum Example}
\begin{figure}[t]
    \centering
    \input{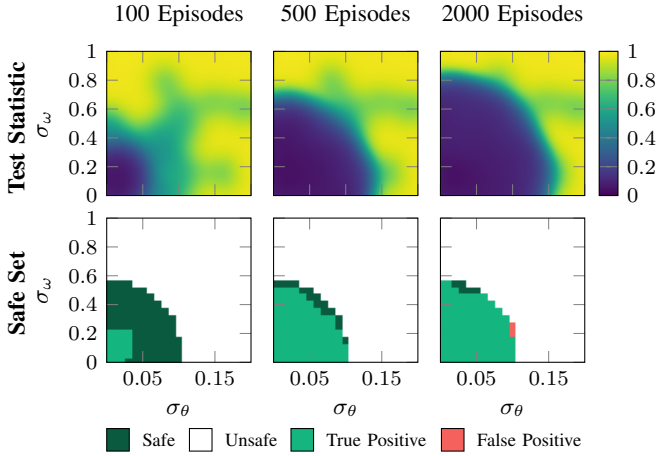}
    \vspace{-3mm}
    \caption{Pendulum smoothing bandit DKWUCB acquisition. \label{fig:pend_smooth_acq}}
    \vspace{-3mm}
\end{figure}
\Cref{fig:pend_smooth_acq} shows the progression of the smoothed DKWUCB algorithm on the pendulum problem using exploration constant $c=1$ and length parameter $\ell=0.02$. The top row shows the smoothed test statistic, and the bottom row shows the safe set estimates. After simulating only \num{500} episodes (less than \SI{0.01}{\percent} of the number of episodes simulated in the na\"ive Monte Carlo method), the algorithm has identified most of the points in the ground truth safe set. After \num{2000} iterations, the safe set estimate contains two false positives. While this result is still below the number of false positives implied by using $\delta=0.95$, we should be careful not to select a length parameter $\ell$ that is too large. The following section describes a principled way to select this parameter.

\subsection{Kernel Learning}\label{sec:kernel_learning}
We propose to use the kernel-smoothed counts not only to select points for evaluation but also to select which points should be included in the safe set. Therefore, we are directly using the kernel function to assess properties related to safety, and it is important that we select a kernel function that will produce an accurate model. We note that we can only achieve the precision implied by the selected confidence $\delta$ if our probabilistic model over the probability of failure is accurate. This section develops a method to learn a kernel from the data we observe. 

Previous work on kernel learning for Gaussian processes has focused on estimating the kernel parameters using a maximum likelihood model and gradient-based optimization techniques \cite{alg4opt}. In this work, we will be more conservative by estimating the  distribution over the kernel parameters conditioned on observed data. In particular, we focus on learning a distribution over the length parameter for the weighted squared exponential kernel in \cref{eq:wsqe_kernel}. Since the correlation between the probability of failure at neighboring points changes in different regions of the performance characteristic space (see \cref{fig:pend_gt}), we estimate a different length parameter distribution for each point $\eta \in \Omega$.

We are specifically interested in estimating a probability distribution over the length parameter at point $\eta$ given the observed data $\mathcal D$, which we will write as $P(\ell_\eta \mid \mathcal D)$. The observed data consists of the number of successes we have observed $\vect p$ at each point and the number of failures observed $\vect q$. Therefore, we can write
\begin{equation}
    P(\ell_\eta \mid \mathcal D) = P(\ell_\eta \mid \vect p, \vect q, p_\eta, N_\eta)
\end{equation}
where $p_\eta$ and $N_\eta = p_\eta + q_\eta$ are quantities derived from $\vect p$ and $\vect q$. The reason for writing the data quantities in this way will become apparent in the next few steps. Next, we apply Bayes' rule to this expression to obtain
\begin{equation}
    \begin{split}
        P(\ell_\eta \mid \vect p, \vect q, p_\eta, N_\eta) & \propto P(p_\eta \mid \ell_\eta, \vect p, \vect q, N_\eta) P(\ell_\eta \mid  \vect p, \vect q, N_\eta) \\
        & = P(p_\eta \mid \ell_\eta, \vect p, \vect q, N_\eta) P(\ell_\eta)
    \end{split}
\end{equation}
where $P(\ell_\eta)$ is a prior distribution that does not depend on $\vect p$, $\vect q$, or $N_\eta$.

Given the length parameters and observed data, we can determine the kernel smoothed counts for point $\eta$ as follows
\begin{equation}\label{eq:est_counts}
    \begin{split}
        \hat p_\eta & = [\mathbf K (\vect \eta, \vect \eta)]_\eta \vect p \\
        \hat q_\eta & = [\mathbf K (\vect \eta, \vect \eta)]_\eta \vect q
    \end{split}
\end{equation}
where $[\mathbf K (\vect \eta, \vect \eta)]_\eta$ is the row in the kernel matrix corresponding to $\eta$. We can then rewrite $P(p_\eta \mid \ell_\eta, \vect p, \vect q, N_\eta)$ and simplify by introducing the probability of success $\theta_\eta$ as follows
\begin{equation}
    \begin{split}
        P(p_\eta \mid \ell_\eta,& \vect p, \vect q, N_\eta) = P(p_\eta \mid \hat p_\eta, \hat q_\eta, N_\eta) \\
        & = \int_{0}^1 P(p_\eta, \theta_\eta \mid \hat p_\eta, \hat q_\eta, N_\eta) d\theta_\eta \\
        & = \int_{0}^1 P(p_\eta \mid \theta_\eta, \hat p_\eta, \hat q_\eta, N_\eta) P(\theta_\eta \mid \hat p_\eta, \hat q_\eta, N_\eta)d\theta_\eta \\
        & = \int_{0}^1 P(p_\eta \mid \theta_\eta, N_\eta) P(\theta_\eta \mid \hat p_\eta, \hat q_\eta)d\theta_\eta \\
        & = \int_{0}^1 \text{Binomial}(p_\eta \mid \theta_\eta, N_\eta) \text{Beta}(\theta_\eta \mid \hat p_\eta, \hat q_\eta)d\theta_\eta
    \end{split}
\end{equation}
The resulting integral can be computed exactly as follows
\begin{equation}\label{eq:p_ell}
    \begin{split}
        P( & p_\eta \mid \ell_\eta, \vect p, \vect q, N_\eta) \\
        & = \frac{\Gamma(\hat a + \hat b)\Gamma(a + b - 1)\Gamma(\hat a + a - 1)\Gamma(\hat b + b - 1)}{\Gamma(\hat a)\Gamma(a)\Gamma(\hat b)\Gamma(b)\Gamma(\hat a + \hat b + a + b - 2)}
    \end{split}
\end{equation}
where $a = p_\eta + 1$, $\hat a = \hat p_\eta + 1$, $b = q_\eta + 1$, $\hat b = \hat q_\eta + 1$, and $\Gamma(\cdot)$ is the gamma function. This expression allows us to compute $P(\ell \mid \eta)$ up to a proportionality constant. To compute the full distribution, we discretize the possible values of $\ell_\eta$, apply \cref{eq:p_ell} to each discrete value, and normalize.

Given a distribution over $\ell_\eta$, we can compute the distribution over $\theta_\eta$ (and therefore implicitly over $P_\text{fail}(\eta)$) as follows
\begin{equation}
    \begin{split}
        P(& \theta_\eta \mid \vect p , \vect q, p_\eta, N_\eta) \\
        & = \int_{\ell_\eta} P(\ell_\eta \mid \theta_\eta, \vect p, \vect q, p_\eta, N_\eta) P(\theta_\eta \mid \vect p, \vect q, p_\eta, N_\eta) d\ell_\eta \\
        & = \int_{\ell_\eta} P(\ell_\eta \mid \theta_\eta, \vect p, \vect q, p_\eta, N_\eta) \text{Beta}(\theta_\eta \mid \hat p_\eta, \hat q_\eta) d\ell_\eta
    \end{split}
\end{equation}
We again approximate this integral by discretizing the possible values for $\theta$ and normalizing. We perform kernel learning in the loop in \cref{alg:safe_set_est}. At each iteration, we update our distribution over the length parameter at each point in $\Omega$ after observing a result from the black-box simulator.

\begin{figure*}[t]
    \centering
    \input{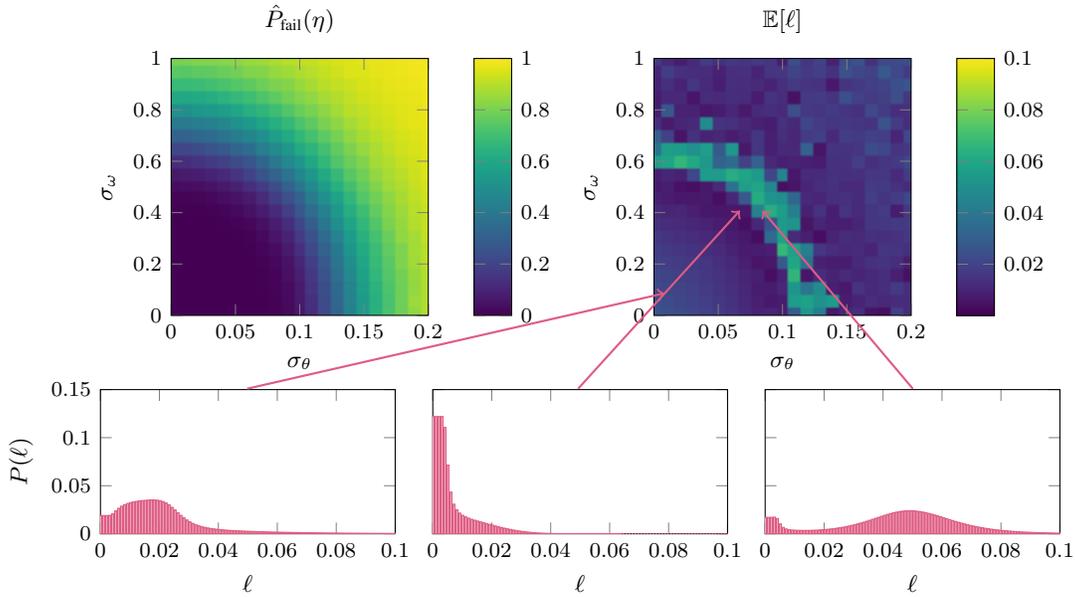}
    \caption{Pendulum kernel learning results. \label{fig:pend_kern_learn}}
\end{figure*}

\paragraph{Kernel Learning Intuition} The underlying assumption we make when we apply kernel smoothing to the observed counts is that the failure probabilities at nearby points are correlated. However, if this assumption is violated, our model will be incorrect and could produce misleading safe sets. By allowing the model to learn a kernel function for each point, we allow it to incorporate a notion of spatial correlation \textit{only if} it is consistent with the data we observe. For points in regions where the failure probability is rapidly changing, we expect the model to estimate a lower length parameter. In contrast, in regions where the failure probability is relatively constant or linear, the model will likely estimate a high length parameter and can benefit from the added efficiency of accounting for this spatial correlation.

\paragraph{Pendulum Example} To estimate a distribution over the length parameter for each point in the pendulum problem, we discretize $\ell$ into \num{100} bins. We then apply \cref{eq:p_ell} at each point $\eta \in \Omega$ using the observed counts after running the DKWUCB algorithm for \num{50000} episodes. \Cref{fig:pend_kern_learn} shows the expectation of each distribution over the length parameter alongside the ground truth probability of failure. The example distributions on the left and in the middle show how the distribution over the length parameter shifts toward smaller values in regions of rapid change. The distribution on the right shows an example of a distribution in the band of high length parameter estimates. In this region, the change in probability of failure is roughly linear, which is well-represented by the smoothed counts.

\section{Aircraft Collision Avoidance Application}
\begin{figure}[t]
    \centering
    \input{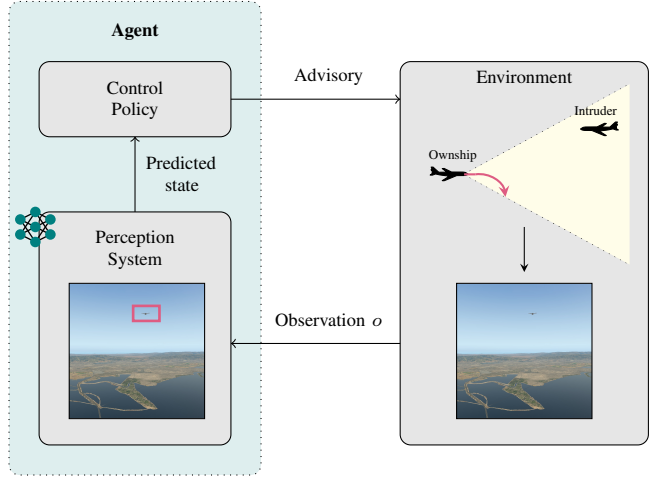}
    \caption{Overview of vision-based aircraft collision avoidance system. \label{fig:daa_overview}}
    \vspace{-5mm}
\end{figure}
Aircraft designers use multiple different sensors and algorithms for surveillance and tracking, resulting in a wide range of perception system performance characteristics. For the purpose of safe design and regulation, it is useful to specify a set of performance characteristics that will result in an acceptable collision risk. We can apply the methods developed in this work to enumerate this set. Specifically, we consider the system shown in \cref{fig:daa_overview}. A set of onboard sensors such as cameras monitor the surrounding sky. The observations from these sensors are passed through a perception system, which produces an estimate of the relative geometry of intruding aircraft. This estimate is then used by a controller to select a collision avoidance maneuver.

Before we run our set estimation algorithm, we must define a representative set of performance characteristics for the perception component of the system shown in \cref{fig:daa_overview}. Key attributes of this system include field of view and detection probability. Detection probability depends strongly on the range between the ownship and intruder, as aircraft that are further away will represent fewer pixels in the image. We develop a model for detection probability based on the performance of the baseline and risk-driven perception neural networks used in \citeauthor{corso2022risk} \cite{corso2022risk}.

\begin{figure}[t]
    \centering
    \begin{tikzpicture}[]
\begin{groupplot}[group style={horizontal sep = 0.75cm, group size=2 by 2}]

\nextgroupplot [
  height = {4.8cm},
  ylabel = {Detection Probability},
  xmin = {0.0},
  xmax = {3000.0},
  ymax = {1.05},
  xlabel = {Range (m)},
  ymin = {0.0},
  width = {4.8cm},
  title = {Baseline}
]

\addplot+[
  ybar, bar width=2.5pt, draw=gray, fill=gray!30, mark=none
] coordinates {
  (50.0, NaN)
  (150.0, 0.2777777777777778)
  (250.0, 1.0)
  (350.0, 0.9891304347826086)
  (450.0, 0.9477611940298507)
  (550.0, 0.8834355828220859)
  (650.0, 0.803921568627451)
  (750.0, 0.7184873949579832)
  (850.0, 0.7396694214876033)
  (950.0, 0.6423076923076924)
  (1050.0, 0.4896551724137931)
  (1150.0, 0.4022140221402214)
  (1250.0, 0.29069767441860467)
  (1350.0, 0.2533783783783784)
  (1450.0, 0.18518518518518517)
  (1550.0, 0.08396946564885496)
  (1650.0, 0.09433962264150944)
  (1750.0, 0.07042253521126761)
  (1850.0, 0.0728476821192053)
  (1950.0, 0.0299625468164794)
  (2050.0, 0.0)
  (2150.0, 0.0)
  (2250.0, 0.0)
  (2350.0, 0.0)
  (2450.0, 0.0071174377224199285)
  (2550.0, 0.0)
  (2650.0, 0.0)
  (2750.0, 0.0)
  (2850.0, 0.0)
  (2950.0, 0.0038910505836575876)
  (3050.0, 0.0)
  (3150.0, 0.0)
  (3250.0, 0.0)
  (3350.0, 0.0)
  (3450.0, 0.0)
  (3550.0, 0.0)
  (3650.0, 0.0)
  (3750.0, 0.0)
  (3850.0, 0.0)
  (3950.0, 0.0)
  (4050.0, 0.0)
  (4150.0, 0.0)
  (4250.0, 0.0)
  (4350.0, 0.0)
  (4450.0, 0.0)
};

\addplot+[
  very thick, teal, mark=none
] coordinates {
  (0.0, 0.0)
  (200.0, 0.0)
  (200.0, 1.0106060606060605)
  (1650.0, 0.0)
  (1650.0, 0.0)
  (3000.0, 0.0)
};

\nextgroupplot [
  height = {4.8cm},
  xmin = {0.0},
  xmax = {3000.0},
  ymax = {1.05},
  xlabel = {Range (m)},
  ymin = {0.0},
  width = {4.8cm},
  title = {Risk-Driven}
]

\addplot+[
  ybar, bar width=2.5pt, draw=gray, fill=gray!30, mark=none
] coordinates {
  (50.0, NaN)
  (150.0, 0.2222222222222222)
  (250.0, 0.9433962264150944)
  (350.0, 0.9456521739130435)
  (450.0, 0.9477611940298507)
  (550.0, 0.8895705521472392)
  (650.0, 0.8627450980392157)
  (750.0, 0.7310924369747899)
  (850.0, 0.8140495867768595)
  (950.0, 0.7461538461538462)
  (1050.0, 0.6482758620689655)
  (1150.0, 0.6346863468634686)
  (1250.0, 0.5736434108527132)
  (1350.0, 0.5641891891891891)
  (1450.0, 0.5037037037037037)
  (1550.0, 0.4389312977099237)
  (1650.0, 0.4490566037735849)
  (1750.0, 0.3732394366197183)
  (1850.0, 0.3543046357615894)
  (1950.0, 0.23595505617977527)
  (2050.0, 0.0076045627376425855)
  (2150.0, 0.0035714285714285713)
  (2250.0, 0.007692307692307693)
  (2350.0, 0.007352941176470588)
  (2450.0, 0.014234875444839857)
  (2550.0, 0.0)
  (2650.0, 0.015503875968992248)
  (2750.0, 0.0)
  (2850.0, 0.007326007326007326)
  (2950.0, 0.011673151750972763)
  (3050.0, 0.007751937984496124)
  (3150.0, 0.011363636363636364)
  (3250.0, 0.0)
  (3350.0, 0.008620689655172414)
  (3450.0, 0.00847457627118644)
  (3550.0, 0.004672897196261682)
  (3650.0, 0.011363636363636364)
  (3750.0, 0.010810810810810811)
  (3850.0, 0.007633587786259542)
  (3950.0, 0.0)
  (4050.0, 0.0)
  (4150.0, 0.0)
  (4250.0, 0.0)
  (4350.0, 0.0)
  (4450.0, 0.0)
};

\addplot+[
  very thick, teal, mark=none
] coordinates {
  (0.0, 0.0)
  (200.0, 0.0)
  (200.0, 0.9384)
  (2000.0, 0.20400000000000007)
  (2000.0, 0.0)
  (3000.0, 0.0)
};

\end{groupplot}

\filldraw[rounded corners, fill=gray!20, draw=black] (0.85, 2.2) rectangle (3.1, 3.1);
\node[anchor=east] at (3.1, 2.8) {\scriptsize True NMACs: 54};
\node[anchor=east] at (3.1, 2.5) {\scriptsize \textcolor{teal}{Model NMACs: 88}};

\filldraw[rounded corners, fill=gray!20, draw=black] (4.82, 2.2) rectangle (7.07, 3.1);
\node[anchor=east] at (7.07, 2.8) {\scriptsize True NMACs: 23};
\node[anchor=east] at (7.07, 2.5) {\scriptsize \textcolor{teal}{Model NMACs: 41}};

\end{tikzpicture}
    \vspace{-3.5mm}
    \caption{Aircraft detection system modeling. \label{fig:detect_model}}
    \vspace{-3mm}
\end{figure}
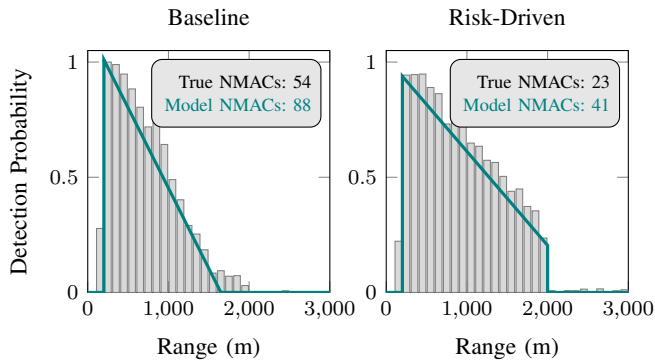

\Cref{fig:detect_model} shows the effect of range on probability of detection for both networks when evaluated on a dataset of images with intruder location sampled uniformly within the ownship field of view. Based on these results, we use the following conservative model of the detection probability
\begin{equation}
    P_\text{detect}(r) = \begin{cases}
        -\frac{y_0}{x_0} r + y_0, & 200 < r < 1000 \\
        0, & \text{else}
    \end{cases}
\end{equation}
where $r$ is the range in meters between the ownship and intruder, and $x_0$ and $y_0$ are the $x$-intercept and $y$-interecept of a line that models the dropoff rate in detection probability. The teal lines in \cref{fig:detect_model} show different parameterizations of this model for each example network.

We produce a conservative model by selecting values for $x_0$ and $y_0$ such that the true detection probabilities always remain at or above the dropoff line. To check that our estimate is conservative, we generated a set of \num{200} aircraft encounters using the encounter model from \citeauthor{corso2022risk} \cite{corso2022risk}. We simulated the true networks and our perception system models on this set and evaluated the number of near mid-air collisions (NMACs). As expected, the conservative perception models resulted in a higher number of NMACs; however, the models still preserved the trends in safety. Future work could study less conservative ways to model the perception system.

\begin{figure}[t]
    \centering
    \input{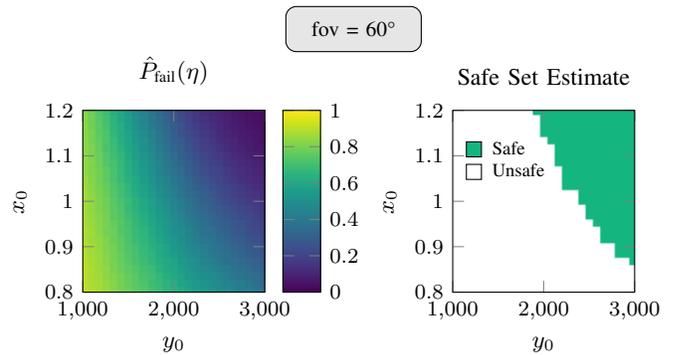}
    \vspace{-3mm}
    \caption{Ground truth results for the aircraft collision avoidance application. \label{fig:gt_daa}}
    \vspace{-4mm}
\end{figure}

In summary, we abstract the perception system using three performance characteristics such that $\eta=[x_0, y_0, h_\text{fov}]$, where $h_\text{fov}$ is the horizontal field of view. We can apply a na\"ive Monte Carlo approach to estimate the safe set given a threhold of $\gamma=0.3$. For this analysis, we discretized $x_0$ using \num{21} points between \num{1000} and \num{3000}, $y_0$ using \num{21} points between \num{0.8} and \num{1.2}, and $h_\text{fov}$ using \num{8} points between \SI{30}{\degree} and \SI{100}{\degree}. We then simulated \num{10000} encounters randomly generated using the model described in \citeauthor{corso2022risk} \cite{corso2022risk} at each point to estimate the probability of NMAC. \Cref{fig:gt_daa} shows the results for a slice of the performance characteristic space in which the horizontal field of view is fixed at \SI{60}{\degree}. We use these results as ground truth to evaluate the performance of our methods.

\section{Experimental results}

\begin{figure}[t]
    \centering
    \input{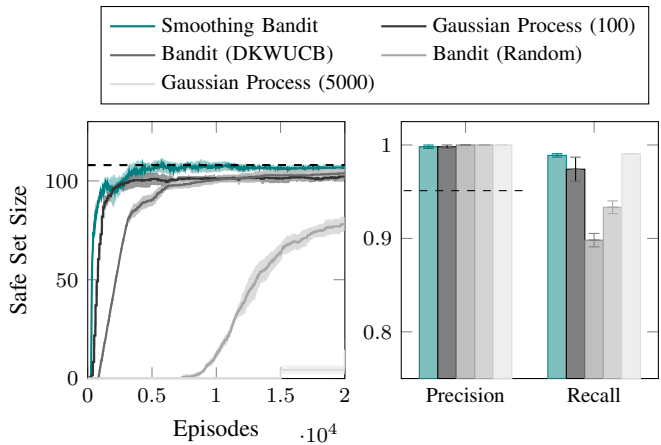}
    \caption{Pendulum results comparison. \label{fig:pend_res}}
\end{figure}

\begin{figure}[t]
    \centering
    \input{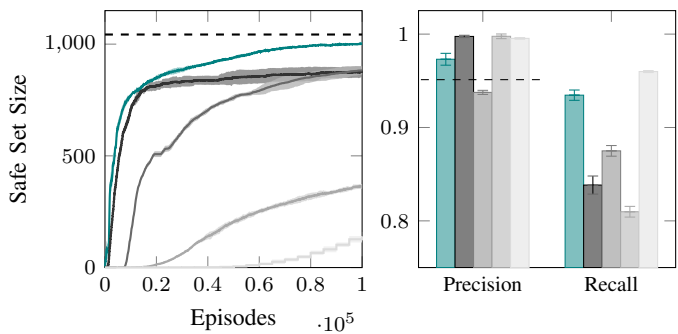}
    \vspace{-3mm}
    \caption{Aircraft collision avoidance results comparison. \label{fig:daa_res}}
\end{figure}

To analyze the efficacy of the proposed smoothing bandit safe-set estimation method, we compare to the baseline methods described in \cref{sec:baselines}. We ran five trials of each method on both the pendulum problem and aircraft collision avoidance problem until convergence and evaluated the models. We evaluate efficiency by analyzing the number of episodes required to enumerate the safe set. Robustly extracting the safe set using confidence $\delta$ should limit the number of false positives so long as our model of the distribution over the probability of failure is accurate. We assess this accuracy by checking the precision and recall of the converged models. The results are shown in \cref{fig:pend_res} and \cref{fig:daa_res}.

The Gaussian process baseline method has hyperparameters for the length parameter of the kernel and the number of simulation episodes per evaluation. We selected a length parameter of $\ell=0.1$ for both problems by performing a grid search; however, we note that this selection method will result in generous performance measures since this grid search would generally be intractable for larger problems. We use two different values for the number of simulation episodes: \num{100} and \num{5000}. A small number of simulation episodes will result in a high variance $\nu$ on evaluated points. Therefore, the Gaussian process with \num{100} episodes is unable to resolve points on the border and biases toward an underestimate of the safe set resulting in a low recall. We can alleviate this bias by simulating \num{5000} episodes per evaluation, but this method results in a large drop in efficiency.

For the threshold bandit baseline, we used both an acquisition function that selects arms at random and the DKWUCB acquisition. Using the DKWUCB acquisition greatly improves efficiency in enumerating the safe set. In both applications, the smoothing bandit with kernel learning technique results in the best efficiency while maintaining a high precision and recall. For our analysis, we selected $\delta=0.95$, which should fix the precision to be above \num{0.95} so long as the probabilistic model is accurate. We denote this threshold with a dashed line on the precision results in \cref{fig:pend_res} and \cref{fig:daa_res}. The smoothing bandit model meets this precision threshold while also efficiently enumerating the safe set, which supports the efficacy of our proposed modeling and kernel learning techniques.

\section{Conclusion}
In this work, we presented a method to translate closed-loop safety properties to safety requirements for perception systems. We showed how to efficiently determine sets of safe performance characteristics given a black-box simulator of the closed-loop system. We presented two baseline methods involving common black-box estimation techniques including level-set estimation using Gaussian processes and threshold bandits. We then proposed a smoothing bandit technique that combines the advantages of each baseline and showed how we can learn from observed data to produce accurate results. The output of this method can be used to inform the design of the perception system. 

\renewcommand*{\bibfont}{\small}
\printbibliography

\end{document}